\documentclass[letterpaper]{article} 
\usepackage{aaai25}  
\usepackage{times}  
\usepackage{helvet}  
\usepackage{courier}  
\usepackage[hyphens]{url}  
\usepackage{graphicx} 
\usepackage{natbib}  
\usepackage{caption} 
\usepackage{algorithm}
\usepackage{algorithmic}
\usepackage{multirow}
\usepackage{makecell}
\usepackage{latexsym}
\usepackage{amssymb}
\usepackage{amsmath}
\usepackage{amsthm}
\usepackage{booktabs}
\usepackage{subcaption} 
\usepackage{newfloat}
\usepackage{listings}
\DeclareCaptionStyle{ruled}{labelfont=normalfont,labelsep=colon,strut=off} 
\floatstyle{ruled}
\newfloat{listing}{tb}{lst}{}
\floatname{listing}{Listing}

\newcommand{\RNum}[1]{\uppercase\expandafter{\romannumeral #1\relax}}
\title{SWEA: Updating Factual Knowledge in Large Language Models via Subject Word Embedding Altering}
\author{
    Xiaopeng Li,
    Shasha Li$^*$,
    Shezheng Song,
    Huijun Liu,
    Bin Ji,
    Xi Wang,
    Jun Ma\thanks{Corresponding Author.},
    Jie Yu$^*$,
    Xiaodong Liu$^*$,
    Jing Wang,
    Weimin Zhang
}
\affiliations{
    National University of Defense Technology\\
    Changsha, Hunan 410073 China\\
 \{xiaopengli, shashali, ssz614, jibin, liuhuijun, wx\_23ndt, yj, majun, liuxiaodong, wangjing\}@nudt.edu.cn, wmzhang104@139.com
}

\begin{document}

\maketitle

\begin{abstract}
The general capabilities of large language models (LLMs) make them the infrastructure for various AI applications, but updating their inner knowledge requires significant resources. Recent model editing is a promising technique for efficiently updating a small amount of knowledge of LLMs and has attracted much attention. In particular, local editing methods, which directly update model parameters, are proven suitable for updating small amounts of knowledge. Local editing methods update weights by computing least squares closed-form solutions and identify edited knowledge by vector-level matching in inference, which achieve promising results. However, these methods still require a lot of time and resources to complete the computation. Moreover, vector-level matching lacks reliability, and such updates disrupt the original organization of the model's parameters.
To address these issues, we propose a detachable and expandable Subject Word Embedding Altering (SWEA) framework, which finds the editing embeddings through token-level matching and adds them to the subject word embeddings in Transformer input. To get these editing embeddings, we propose optimizing then suppressing fusion method, which first optimizes learnable embedding vectors for the editing target and then suppresses the Knowledge Embedding Dimensions (KEDs) to obtain final editing embeddings.
We thus propose SWEA$\oplus$OS method for editing factual knowledge in LLMs. We demonstrate the overall state-of-the-art (SOTA) performance of SWEA$\oplus$OS on the \textsc{CounterFact} and zsRE datasets. To further validate the reasoning ability of SWEA$\oplus$OS in editing knowledge, we evaluate it on the more complex \textsc{RippleEdits} benchmark. The results demonstrate that SWEA$\oplus$OS possesses SOTA reasoning ability.
\end{abstract}

%

\section{Introduction}
Large language models (LLMs), with their rich reserve of pre-trained knowledge, play a pivotal role in the current AI landscape \cite{li2023multimodal,zhao2023survey}. The knowledge pre-trained in LLMs is solidified in their parameters, meaning that any outdated or incorrect knowledge within the LLMs can only be updated through parameter updates. However, given that the training of LLMs relies heavily on GPUs and consumes a significant amount of electricity, retraining to update even small amounts of information can be costly. Consequently, researchers have started to explore model editing methods \cite{yao-etal-2023-editing,Meng2022Locating,mitchell2022fast,zhang2024comprehensive,wang2024wise,tian2024instructedit} aiming to update a small amount of knowledge of LLMs more efficiently.

 \begin{figure}[t]
  \centering
  \includegraphics[scale=0.5]{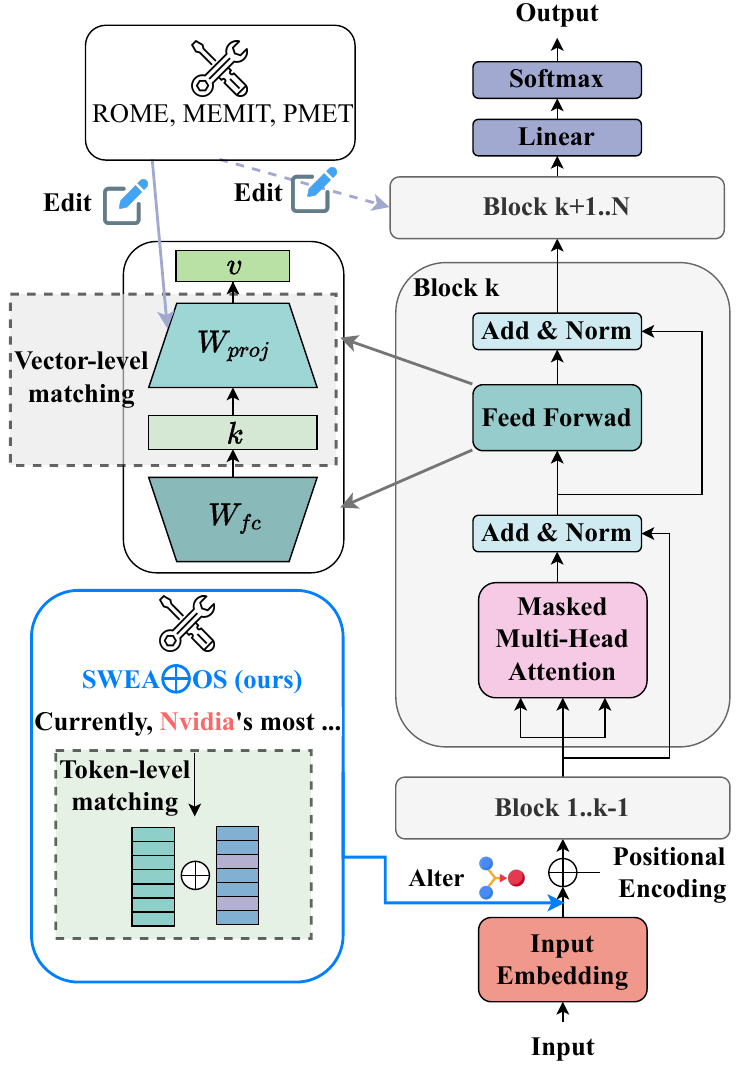}
  \caption{Difference between our method and existing local editing methods. Our method focuses on altering the word embedding for the input via token-level matching, while existing local editing methods edit Feed Forward Network (FFN) and identify editing knowledge by vector-level matching. Mismatching is more likely to occur in vector-level matching, which leads to erroneous recognition of editing knowledge.}\label{fig:SWEA_loc}
\end{figure}
The purpose of model editing is to insert, update, and delete target knowledge while avoiding editing non-target knowledge to preserve the original capabilities of LLMs. 
Current editing methods mainly edit model through three approaches \cite{wang2023knowledge}: adding additional modules \cite{huang2023transformer,dong2022calibrating,hartvigsen2022aging}, global optimization \cite{zhu2020modifying,mitchell2022fast}, and local editing \cite{Meng2022Locating,meng2022massediting,li2023pmet}. Methods of adding additional modules involves incorporating adapters within or external to the LLMs for storing edited instances, which increases the inference load. In contrast, global optimization and local editing methods write the editing information into the model weights, maintaining the same inference cost as the original model. However, using global optimization methods for model editing is prone to overfitting \cite{meng2022massediting} because model editing often only requires updating a small amount of knowledge. Local editing methods view the model editing as a least squares problem, which is more suitable for updating a small amount of knowledge. Therefore, in this paper, we focus on local editing methods.

Local editing methods first select the critical layers that store knowledge \cite{Meng2022Locating}, optimize the knowledge representation with the editing knowledge as the objective, then calculate the keys of the editing knowledge and the original knowledge, and finally update the weights of critical layers by solving the least squares problem. These methods have achieved remarkable results in model editing tasks. However, they still exsit three issues: \textbf{\textit{(1) Lack of efficiency:}} these methods need to spend a lot of time and resources to compute all the vectors needed to solve the least squares problem \cite{Meng2022Locating,meng2022massediting,li2023pmet}; \textbf{\textit{(2) Lack of reliability:}} in local editing methods, we observe that even when all the target knowledge representations are already aligned with the editing goal, their editing success rate is still far from expectation. Meanwhile, LLMs edited by these methods are prone to misidentifying unedited knowledge as edited knowledge, reducing the usability of edited LLMs. This might be due to the fact that using vector-level matching to identify the editing knowledge in the updated weights is not entirely reliable, since vector-level matching struggles to distinguish between two very similar vectors \cite{gionis1999similarity,zhang2023bort}; \textbf{\textit{(3) Lack of protection:}} 
due to the high complexity and incomplete transparency of LLMs themselves, 
exactly updating their weights perfectly by solving the least squares problem is challenging. Consequently, the original organization of the edited model's parameters is disrupted \cite{li2023unveiling}, thereby affecting the general applications of LLMs \cite{gu2024model}. 

In view of these issues, \textbf{\textit{(1)}} we propose a novel model editing method, SWEA$\oplus$OS, which alters subject word embedding by adding it with editing embeddings obtained by \textbf{O}optimizing then \textbf{S}uppressing (OS) fusion method in \textbf{S}ubject \textbf{W}ord \textbf{E}mbedding \textbf{A}ltering (SWEA) framework. SWEA$\oplus$OS only requires computing editing embeddings, therefore it is more efficient. The difference between SWEA$\oplus$OS and existing local editing methods is illustrated in Figure \ref{fig:SWEA_loc}. \textbf{\textit{(2)}} The SWEA framework identifies editing knowledge instances through token-level matching that is more reliable than vector-level matching because it is sensitive to even single-character changes. The OS fusion method get the editing embeddings through: a) optimizing learnable embedding vectors to achieve editing objectives, b) suppressing the subject's Knowledge Embedding Dimensions (KEDs) which are special dimensions related to specific knowledge in word embeddings. The suppressing step is designed to mitigate the influence of the subject's KEDs on the expression of new knowledge. \textbf{\textit{(3)}} Unlike local editing methods that directly modify weights, the SWEA framework is detachable and embedded into the embedding layer of LLMs, which protects the original weights of LLMs. It is also expandable, which can be combined with different fusion methods for model editing. In addition, the SWEA framework edits knowledge by altering the subject word embedding, which ensures the same inference load as the original model. 

We demonstrate our method is both efficient and effective in GPT-J (6B) \cite{wang2021gpt} and Llama-2 (7B) \cite{touvron2023llama} across two datasets and one benchmark. In detail, comparative experiments on GPT-J and Llama-2 show that the SWEA$\oplus$OS method demonstrates overall SOTA performance. On the \textsc{CounterFact} dataset, SWEA$\oplus$OS increases the Score by 5.8\% on GPT-J and 7.7\% on Llama-2 compared to the most advanced baseline. The SWEA$\oplus$OS method also shows the best reasoning performance on the RippleEdits benchmark \cite{cohen2023evaluating}, indicating that the knowledge edited by the SWEA
OS method has stronger consistency.

 \begin{figure*}[t]
  \centering
  \includegraphics[scale=0.62, trim={0 0 0 0},clip]{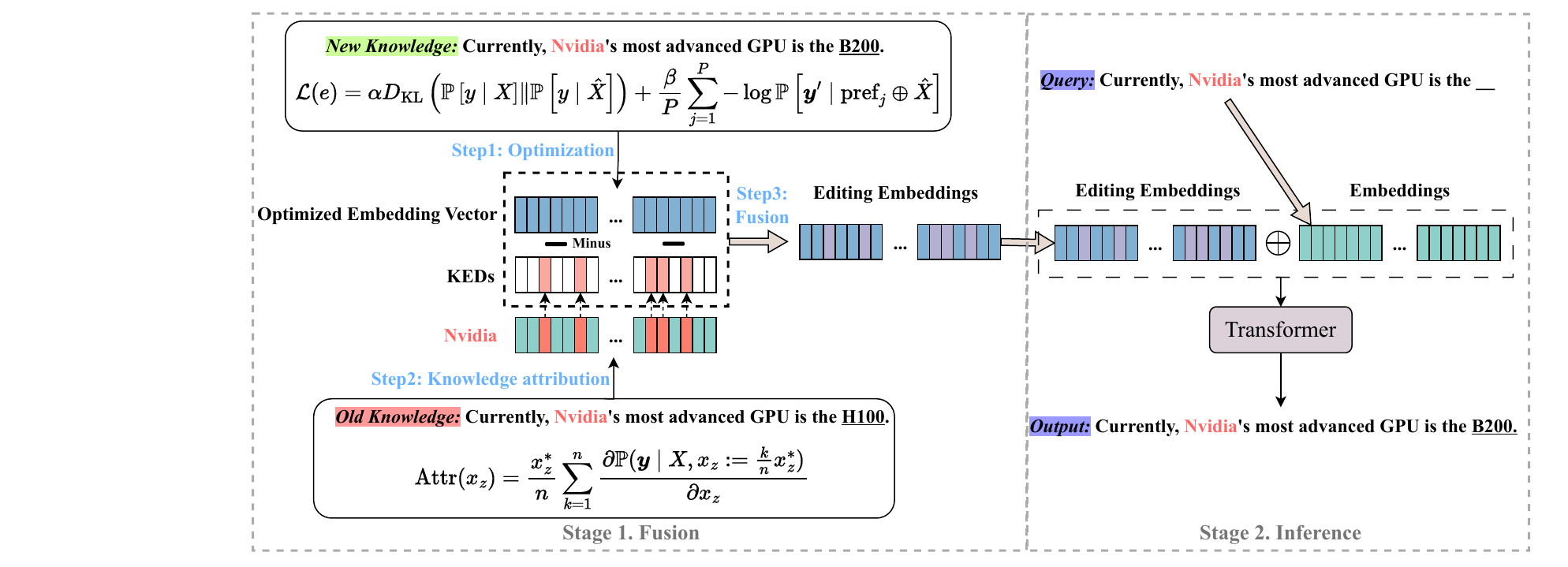}
  \caption{Overview of SWEA$\oplus$OS. In fusion stage, we first optimize a learnable embedding vector for target knowledge ``Currently, Nvidia's most advanced GPU is the \underline{B200}.'' Second, using knowledge attribution method, we find the KEDs of `Nvidia' regarding ``its most advanced GPU''. Finally, we fuse the optimized embedding vector with these KEDs subtracted to obtain the editing embeddings. In inference stage, we add these editing embeddings to the embedding of the subject `Nvidia' for inference.}\label{fig:SWEAOS:overview}
\end{figure*}

Our contributions to the work are as follows:
\begin{itemize}
\item We propose a detachable and expandable SWEA framework, which can be combined with different fusion methods for model editing and ensures the same inference cost as the original model.
\item We introduce the OS fusion method. It optimizes learnable embedding vectors for editing targets and then suppresses KEDs of subject to alleviate the impact of KEDs of subject word embeddings on editing effects.
\item Combing the OS fusion method with SWEA, we propose SWEA$\oplus$OS for editing factual knowledge in LLMs. We demonstrate the overall superior performance of SWEA$\oplus$OS on \textsc{CounterFact} and zsRE datasets and a more complex \textsc{RippleEdits} benchmark.
\end{itemize}
\section{Related Work}
\subsection{Model Editing} 
Model editing is currently an emerging research hotspot, with various model editing methods and benchmarks being proposed successively \cite{yao-etal-2023-editing,zhang2024comprehensive,wang2023knowledge,deng2024unke,wang2024wise,li2024consecutive}. The model editing task was first proposed in \cite{zhu2020modifying}, where they proposed a constrained fine-tuning method for this task, which imposes a constraint on the fine-tuning weights to reduce interference with original knowledge. Unlike constrained fine-tuning, recent methods utilize meta-learning to update weights \cite{de-cao-etal-2021-editing,mitchell2022fast,tan2023massive}. These methods train a hypernetwork that indirectly updates weights using global gradient information. However, since the model editing task aims to correct a small portion of errors within the model's internal memories, the data for model editing is usually few, making methods that update weights using global gradients prone to overfitting. Some methods also add additional modules to perform model editing. They usually add a smaller model outside LLMs \cite{huang2023transformer}, embed an adapter within LLMs \cite{hartvigsen2022aging}, or add editing information to the input for model editing. But these methods not only increase the inference burden but also add to the complexity of the system.

In contrast, local editing methods, from the perspective of interpretability, directly update the Feed Forward Network (FFN) of key-value memories \cite{geva-etal-2021-transformer} using a closed-form solution of least squares \cite{Meng2022Locating,meng2022massediting,li2023pmet}, which is less prone to overfitting and more lightweight. However, these approaches still require a lot of time and resources to update weights, the vector-level matching in model weights is not always reliable, and recent works find that these approaches can cause irreversible damage to the model's generalization capability due to the updating of model weights \cite{gu2024model}. Unlike existing local editing methods, SWEA$\oplus$OS alters the subject word embeddings in the input through token-level matching through editing embeddings obtained by OS fusion method, making it more efficient and reliable.  
\subsection{Explanation of Word Embedding}
Word embedding is a fundamental component in LLMs for processing natural language, where each token typically corresponds to a high-dimensional dense vector. Researchers have sought to understand the interpretable concepts associated with these dense vectors by categorizing the dimensions into specific concepts \cite{csenel2018semantic,Balogh_Berend_2020} or by projecting word embeddings into more interpretable vector spaces \cite{park-etal-2017-rotated,simhi2023interpreting}. \cite{csenel2018semantic} introduced the SEMCAT dataset and used statistical methods to classify different dimensions of word embedding into 110 semantic categories. \citep{Balogh_Berend_2020} assigned common-sense concepts to each dimension of word embedding. \citep{simhi2023interpreting} mapped word embedding to a concept space understandable by humans using vector similarity. These works enhance our understanding of word embedding in terms of semantic concepts. However, they did not discuss the relationship between the dimensions of word embedding and factual knowledge. In contrast to their methods, we use knowledge attribution methods to identify the corresponding knowledge embedding dimensions (KEDs) for subject-specific facts in word embedding dimensions and suppress these KEDs to improve editing effects.
\section{Preliminaries}
\subsection{Language Modeling}
LLMs process discrete text by firstly embedding it into continuous vectors. After processing by Transformers, a probability distribution on the vocabulary is finally obtained under the action of the Softmax function. Formally, a discrete text $\mathcal{T}$ is first converted into token ids $T=\text{tok}(\mathcal{T})$ by the tokenizer. Next, the embedding layer $E$ of the LLMs maps each token id of $T$ to a vector $x\in R^{1\times h}$, where $h$ is the dimension size of the Transformer's hidden layer. Assuming the length of the token ids is $l$, then $E(T) = X \in R^{l\times h}$ where $X = [x_1, x_2, ..., x_l]$ is the continuous vector of text $\mathcal{T}$. Then the Transformers of the LLMs process $X$ layer by layer, finally obtaining a probability distribution on the vocab:
\begin{equation}\label{equ:SWEA_tansformer}
\mathbb{P}(X) = \text{Softmax}(\text{Transformers}(X))
\end{equation}
\subsection{Model Editing Task}
Prior work expresses factual knowledge as a triple $(s, r, o)$ \cite{meng2022massediting}, where $s$ denotes the subject, $r$ denotes the relation, and $o$ denotes the object. The purpose of model editing is:
\begin{equation}
    (s, r, o) \rightarrow (s,r,o')
\end{equation}where $o'$ is another object different from $o$. At the same time, model editing should protect other knowledge not being changed.
For convenience, we express the factual knowledge with a pair $(\mathcal{T},\mathcal{Y})$, where $\mathcal{T}$ is a sentence composed of $s$ and $r$, and $\mathcal{Y}$ is $o$ which is the continuation of the above sentence. Then the model editing task can be formally expressed as:
\begin{equation}
     (\mathcal{T},\mathcal{Y})\rightarrow  (\mathcal{T},\mathcal{Y}')
\end{equation}

\textbf{Batch Editing} means editing $n>1$ factual knowledge at the same time during a single run of the editing method:
\begin{equation}
     \sum_1^{n}(\mathcal{T},\mathcal{Y})\rightarrow  \sum_1^{n}(\mathcal{T},\mathcal{Y}')
\end{equation}

\textbf{Sequential Editing} means carrying out multiple consecutive edits on a single model:
\begin{equation}
     (\mathcal{T},\mathcal{Y})\rightarrow (\mathcal{T},\mathcal{Y}')\rightarrow ... \rightarrow (\mathcal{T},\mathcal{Y}^*)
\end{equation}

\textbf{Sequential Batch Editing} means performing multiple consecutive batch edits on a single model:
\begin{equation}
     \sum_1^{n}(\mathcal{T},\mathcal{Y})\rightarrow  \sum_1^{n}(\mathcal{T},\mathcal{Y}') \rightarrow ... \rightarrow \sum_1^{n}(\mathcal{T},\mathcal{Y}^*)
\end{equation}
\section{Methodology}
In this section, we explain what is the SWEA framework and how our proposed SWEA$\oplus$OS method is used to update the factual knowledge of LLMs. SWEA$\oplus$OS consists of two stages: (1) Fusion: we use the OS fusion method to compute the editing embeddings needed to update the factual knowledge for the subject; (2) Inference: in the SWEA framework, the input embedding is altered with the matched editing embeddings to obtain the final input embeddings. We detail these two stages in the subsections below, an overview of SWEA$\oplus$OS is shown in Figure \ref{fig:SWEAOS:overview}.
\subsection{Optimizing then Suppressing Fusion Method}\label{sec:ots}
Word embeddings are dense continuous vectors \cite{zhao2023survey}. Some works show that their different dimensions contain specific information \cite{li2016understanding,csenel2018semantic}. Motivated by these, we assume that certain dimensions of word embeddings of a subject correspond to specific factual knowledge about the subject in LLMs. For convenience, we name these dimensions as knowledge embedding dimensions (KEDs). For example, the dimensions $(26, 123, 336, 1024)$ of the word embedding of the subject ``Nvidia'' are KEDs that correspond to the factual knowledge ``Nvidia was founded by Jensen Huang.'' Under this assumption, we aim to alter KEDs of the subject to control the factual knowledge about the subject in LLMs.

Due to word embeddings not being fully explained, directly altering KEDs to update factual knowledge is very difficult. We propose appending learnable embedding vectors to the subject's word embeddings and optimizing these vectors to get optimized embeddings related to the editing target. During inference, simply adding the optimized embedding vectors to the subject's word embeddings can update factual knowledge. However, since the KEDs of the subject's word embeddings corresponding to factual knowledge still work, this may affect the knowledge expression of the optimized embedding vectors, leading to a decrease in editing effects. We thus suppress the KEDs of the original subject's word embeddings.
Therefore, we propose the optimizing then suppressing fusion method, which first optimizes learnable embedding vectors to achieve editing objectives, then suppresses the KEDs of the original subject's word embeddings.

Formally, suppose $X$ is the text embedding of $\mathcal{T}$; $\boldsymbol{y}$ and $\boldsymbol{y}'$ are all tokens of $\mathcal{Y}$ and $\mathcal{Y}'$ respectively. To change the factual knowledge of the model from $(\mathcal{T},\mathcal{Y})$ to $(\mathcal{T},\mathcal{Y}')$, inspired by previous work \cite{meng2022massediting,li2023pmet}, we add learnable embedding vectors $e$ to the representation of the subject $S$ in $X$ to get $\hat{X}$, and use the following loss function to optimize and maximize the probability of $\boldsymbol{y}'$:
\begin{equation}\label{equ:SWEAOS_loss}
  \begin{aligned}
\mathcal{L}(e)&=\alpha D_{\text{KL}}\left(\mathbb{P}\left[y \mid X  \right]  \lVert \mathbb{P}[y \mid \hat{X} ]  \right)+\\
 & \frac{\beta}{P} \sum_{j=1}^P -\log\mathbb{P}\left[\boldsymbol{y}'\mid {\text{pref}}_j \oplus \hat{X}\right]
\end{aligned}
\end{equation}

Here $D_{\text{KL}}$ is the KL divergence used to constrain the probability distribution after adding the learnable embedding vector $e$; to enhance the generalization of the learnable embedding vectors $e$, we prepend $P$ prefixes (i.e., pref$_j$) generated by the model to $\hat{X}$, where $\oplus$ indicates the concatenation operation; $\alpha$ and $\beta$ are two hyperparameters used to regulate the strength between preserving original knowledge and learning new knowledge during the optimization.

We use the knowledge attribution method \cite{dai2021knowledge} to find the KEDs of subject $S$. Let $x_z$ represent any one embedding vector in $x^S  = [x^S_s,...,x^S_e] \in R^{|S|\times h}$, the knowledge attribution of the embedding can be formally expressed as:
\begin{equation}\label{equ:SWEAOS:ka}
  \text{Attr}(x_z) = \frac{x^*_z}{n}\sum_{k=1}^{n}\frac{\partial \mathbb{P}(\boldsymbol{y}\mid X, x_z := \frac{k}{n}x^*_z) }{\partial x_z}
\end{equation}

Here, $x^*_z$ represents the original value of the embedding vector; n is the number of steps for the Riemann integration, and we follow \cite{dai2021knowledge} and set $n = 20$; $\mathbb{P}(\boldsymbol{y}\mid X, x_z  := \frac{k}{n}x^*_z)$ represents the probability of the model generating $\boldsymbol{y}$ after replacing $x_z$ with $\frac{k}{n}x^*_z$. After obtaining the attribution scores of all embedding dimensions of the subject $S$, we retain those embedding dimensions that exceed $t$ times the maximum attribution score as the KEDs $K_D$. Finally, we subtract $\gamma$ times the value of the original embedding vectors $x^S$ corresponding to $K_D$ from the optimized embedding vector $e$ to obtain the final editing embeddings $e^S$:
\begin{equation}\label{equ:SWEAOS:final_es}
  e^S = e - \gamma \mathbb{O}_{\setminus K_D} \odot x^S
\end{equation}where $\mathbb{O}_{\setminus K_D}$ represents a vector with all positions as 0 except for the positions included in $K_D$ which are 1; $\odot$ denotes element-wise multiplication.
\begin{table*}[t]
  \centering
   \begin{tabular}{lrrrrrr}
    \toprule
        {\textbf{Editor}} & \multicolumn{1}{c}{\textbf{Score}} & \multicolumn{1}{c}{\textbf{Efficacy}} & \multicolumn{1}{c}{\textbf{Generalization}} & \multicolumn{1}{c}{\textbf{Specificity}} & \multicolumn{1}{c}{\textbf{Fluency}} & \multicolumn{1}{c}{\textbf{Consistency}} \\
        \midrule
GPT-J & 22.4 & 15.2 (0.7) & 17.7 (0.6) & 83.5 (0.5)  & 622.4 (0.3) & 29.4 (0.2)\\\midrule
FT-W & 67.6 & 99.4 (0.1) & 77.0 (0.7) & 46.9 (0.6) & 293.9 (2.4) & 15.9 (0.3)\\
MEND & 23.1 & 15.7 (0.7) & 18.5 (0.7) & \textbf{83.0} (0.5)  & 618.4 (0.3) & 31.1 (0.2)\\
ROME & 50.3 & 50.2 (1.0) & 50.4 (0.8)  & 50.2 (0.6)  & 589.6 (0.5) & 3.3 (0.0)\\
MEMIT & 85.8 & 98.9 (0.2)  & 88.6 (0.5)  & 73.7 (0.5)  & \underline{619.9} (0.3) & 40.1 (0.2)\\
GRACE & 26.7 & 30.6 (0.9)  &17.3 (0.6)   & \textbf{83.0} (0.5)&618.1 (0.3) &29.3 (0.2) \\
PMET & \underline{86.2} & \underline{99.5} (0.1)  & 92.8 (0.4)  & 71.4 (0.5)  & \textbf{620.0} (0.3) & \underline{40.6} (0.2)\\
SWEA$\oplus$OS & \textbf{91.2} &\textbf{99.6} (0.1)   & \textbf{98.3} (0.2)  & \underline{79.0} (0.5)  & 609.5 (0.7) & \textbf{42.3} (0.2)\\
\toprule
Llama-2 & 20.5 & 14.8 (0.7) & 15.0 (0.6) & 82.4 (0.5)  & 604.3 (0.3) & 25.4 (0.2)\\\midrule
FT-W & 65.4 & \textbf{99.8 (0.1)} & 84.9 (0.6) & {41.5 (0.7)} & {546.9 (0.2)} & {20.0 (0.1)}\\
ROME & 50.5 & 51.3 (1.0) & 50.0 (0.8)  & 50.2 (0.6)  & 488.1 (0.2) & 2.6 (0.0)\\
MEMIT & 69.6 & 81.5 (0.8)  & 55.4 (0.8)  & 78.3 (0.5)  & \underline{602.9} (0.2) & 27.8 (0.2)\\
GRACE & 29.2 & 29.8 (0.9)  & 15.0 (0.6)  & \textbf{82.2} (0.5)&\textbf{605.2} (0.3) &25.3 (0.2) \\
PMET & \underline{83.2} & {97.1} (0.3)  & \underline{87.8} (0.5)  & 69.5 (0.6)  & 599.4 (0.3) & \underline{34.7} (0.2)\\
SWEA$\oplus$OS& \textbf{89.6} & \underline{98.4} (0.2)   & \textbf{93.5} (0.4)   & \underline{79.3} (0.5)  & 600.5 (0.5) & \textbf{35.0} (0.2)\\
\bottomrule
    \end{tabular}
  \caption{Results of 10,000 edits on GPT-J and Llama-2 on the \textsc{CounterFact} dataset. Within the parentheses is the 95\% confidence interval.}\label{tab:SWEAOS:mcf}
\end{table*} 
\subsection{Subject Word Embedding Altering Framework}\label{sec:SWEA}
Subject Word Embedding Altering (SWEA) framework merges the editing embeddings $e^{S} = [e^S_s,...,e^S_e]\in R^{|S|\times h}$ about the subject ${S}$ with the subject embedding $x^S  = [x^S_s,...,x^S_e] \in R^{|S|\times h}$ from the input text embedding $X\in R^{l\times h}$. Here, $|S|$ is the token length of the subject, and $x^S_s$ and $x^S_e$ represent the first and last token of the subject in the input $X$, respectively. Therefore, in SWEA, the final input used by the model for inference is:
\begin{equation}\label{equ:SWEA_input}
  X = [x_0,...,x^S_s+e^S_s,...,x^S_e+e^S_e,...,x_l]
\end{equation}

Given that each subject's token ids are unique, we use these token ids as keys to index the editing embeddings. Specifically, after obtaining the editing embedding $e^{S}$ of the subject $S$, we cache $e^{S}$ using the token ids as key. For $S$ with only one token id, we use this id as the key directly, and for $S$ with multiple token ids, we concatenate the token ids of $S$ using '\_'. For example, the token ids of the subject `San Francisco' are $[2986,6033]$, so its key is `2986\_6033'. For convenience, we currently adopt the file caching method, which can be easily extended to a vector database. SWEA can easily implement batch editing, it can obtain $e^{S}$ for multiple subjects and then cache these $e^{S}$ in editing embeddings collection $\mathcal{E}$. SWEA can also implement sequential editing and sequential batch editing. It caches past editing requests and recomputing $e^{S}$ for the subjects when the editing requests are updated. During the inference stage, we carry out the longest continuous matching for the continuous combination of the token ids of each input and the keys in $\mathcal{E}$, and add successful matched caches to matched tokens's embeddings using \eqref{equ:SWEA_input}. The token-level matching and embedding altering algorithm of the above process can be found in Appendix A (See \cite{li2024sweaupdatingfactualknowledge}). Note that some subjects may have aliases. Currently, we are primarily focused on introducing a new way for model editing, so SWEA currently only considers cases where the subject is unique. However, the SWEA framework can easily be adapted to include an alias list for each subject to identify them.
\section{Experiments}
\subsection{Experimental Setup}
\subsubsection{Datasets and Large Language Models}
We conducted edits on GPT-J (6B) \cite{wang2021gpt} and Llama-2 (7B) \cite{touvron2023llama} on two datasets, \textsc{CounterFact} \cite{Meng2022Locating}, zsRE \cite{de-cao-etal-2021-editing,mitchell2022fast} and the \textsc{RippleEdits} benchmark \cite{cohen2023evaluating}. All metrics of the above datasets and benchmark are described in Appendix B. \textsc{CounterFact} dataset is a completion task, which contains a total of 21,919 counterfactual editing instances. MEMIT \cite{meng2022massediting} filtered out the counter-logical fact editing in this dataset. To ensure the same experimental setting, we also only use the filtered 20,877 editing instances. zsRE dataset is a QA task, for which we use 10,000 editing instances extracted from \cite{Meng2022Locating} to conduct editing. RippleEdits is a benchmark for testing the multi-hop reasoning ability of post-edit models, including \textsc{RECENT}, \textsc{RANDOM} and \textsc{POPULAR} subsets. \textsc{RECENT} mainly evaluates the ability of the model's editing method to insert knowledge, while the latter two mainly evaluate the ability to edit knowledge. Since we currently only focus on updating the knowledge of the model, we only use the two subsets of rippleEdits, \textsc{RANDOM} and \textsc{POPULAR}.
\subsubsection{Baselines}
We compared SWEA$\oplus$OS with the global optimization method Constrained Fine-Tuning (FT+W) \cite{zhu2020modifying}, MEND \cite{mitchell2022fast}, MALMEN \cite{tanmassive}, adding additional modules method GRACE \cite{hartvigsen2022aging}, and the local editing methods ROME \cite{Meng2022Locating}, MEMIT \cite{meng2022massediting}, PMET \cite{li2023pmet} on the \textsc{CounterFact} and zsRE datasets. On the \textsc{RANDOM} and \textsc{POPULAR} subsets of rippleEdits, we compared with local editing methods ROME, MEMIT. Experimental details can be found in Appendix C.
\begin{figure}[t]
    \centering
\includegraphics[width=1\linewidth]{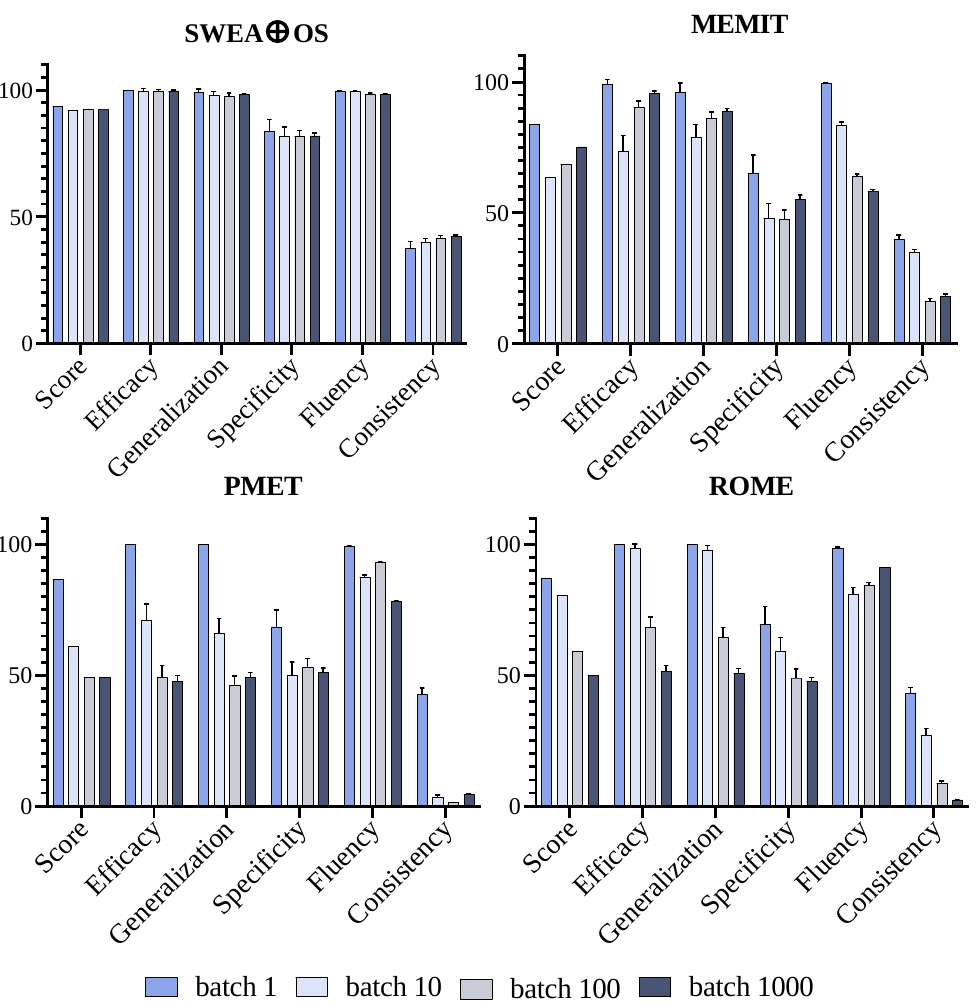}
    \caption{Results of sequential batch editing of SWEA$\oplus$OS, PMET, MEMIT, and ROME. 
    To better display the results, we divide the fluency by the original fluency (i.e., 622.4) and then multiply by 100 to make it fall between 0 and 100.
    }
    \label{fig:SWEAOS:seq_edit}
\end{figure}
\subsection{Experiments on \textsc{CounterFact} and zsRE Datasets}
We first test the batch editing performance on the \textsc{CounterFact} and zsRE datasets. We then test the scaling-up editing performance on the \textsc{CounterFact} dataset. Considering that sequential editing is a subset of sequential batch editing, we perform sequential batch editing directly on the \textsc{CounterFact} dataset. We also compare the execution time of the SWEA$\oplus$OS method with that of baselines and analysis the inference latency introduced by the SWEA framework in Appendix D. 
\begin{table}[t]
  \centering
  \resizebox{.47\textwidth}{!}{
   \begin{tabular}{lrrrr}
   \toprule
\textbf{Editor}&\textbf{Score}&\textbf{Efficacy}&\textbf{Generalization}&\textbf{Specificity}\\
\hline
GPT-J &26.0& 26.4 ($\pm$0.6) & 25.3 ($\pm$0.5) & 26.8 ($\pm$0.5) \\\midrule
FT-W &14.3& 57.9 ($\pm$0.7) & 56.8 ($\pm$0.7) & 5.7 ($\pm$0.5)\\
MEND &20.0&  19.4 ($\pm$0.5) & 18.6 ($\pm$0.5) & 22.4 ($\pm$0.5)\\
MALMEN &37.3&76.1($\pm$0.7)&72.3($\pm$0.7)&18.6($\pm$0.4)\\
ROME &1.1&  9.2 ($\pm$0.8) & 7.9 ($\pm$0.8) & 0.4 ($\pm$0.2)\\
MEMIT &\underline{50.2}&  \underline{92.7} ($\pm$0.3) & \underline{86.7} ($\pm$0.5) & \underline{26.7} ($\pm$0.5)\\
GRACE &31.3&47.8 ($\pm$0.6)&26.5 ($\pm$0.5)& \textbf{26.8} ($\pm$0.5)\\
PMET &47.6&  86.4 ($\pm$0.4) & 81.5 ($\pm$0.5) & 25.5 ($\pm$0.3)\\
SWEA$\oplus$OS &\textbf{51.0}&  \textbf{96.0} ($\pm$0.3) & \textbf{89.7} ($\pm$0.2) & \textbf{26.8} ($\pm$0.2)\\
\toprule
Llama-2 &11.9& 11.5($\pm$0.3) & 11.1 ($\pm$0.3) & 13.3 ($\pm$0.2) \\\midrule
FT-W &11.7& 13.8 ($\pm$0.6) & 13.1 ($\pm$0.5) & {9.2} ($\pm$0.4)\\
ROME &4.3&  3.9 ($\pm$0.8) & 3.7 ($\pm$0.8) & 5.8 ($\pm$0.3)\\
MEMIT &23.1&  45.6 ($\pm$0.4) & 40.9 ($\pm$0.5) & 12.0 ($\pm$0.5)\\
GRACE &14.9&23.7 ($\pm$0.4)&11.8 ($\pm$0.6)&\textbf{13.3} ($\pm$0.5)\\
PMET &\underline{23.9}&  \underline{48.1} ($\pm$0.3) & \textbf{45.0} ($\pm$0.4) & \underline{12.1} ($\pm$0.3)\\
SWEA$\oplus$OS &\textbf{25.5}&  \textbf{50.7 }($\pm$0.4) & \underline{44.0} ($\pm$0.3) &\textbf{ 13.3 }($\pm$0.3)\\
    \bottomrule
    \end{tabular}}
      \caption{Results of 10,000 edits on GPT-J and Llama-2 on the zsRE dataset. To ensure a fair comparison, we reproduced baselines except MEND (non-reproducible) under this setting.}\label{tab:SWEAOS:zsre}
\end{table}
\subsubsection{Batch Editing Results}
The results of editing GPT-J and Llama-2 on the \textsc{CounterFact} and zsRE datasets are presented in Tables \ref{tab:SWEAOS:mcf} and \ref{tab:SWEAOS:zsre}, respectively. SWEA$\oplus$OS achieved the overall best results. Whether on the \textsc{CounterFact} or zsRE datasets, Efficacy, Generalization, Specificity, and Consistency of SWEA$\oplus$OS shows substantial improvement over previous model editing methods. This indicates that SWEA$\oplus$OS is a very effective editing method. FT-W has a very high editing success rate, but the generalizability of the edited knowledge is poor, and the model's generative capability is severely compromised due to overfitting. When editing the GPT-J model, although MEND and GRACE exhibit the best specificity, their poor generalization affects their overall performance. When editing the Llama-2 model, GRACE shows similar results. Overall, we have demonstrated through this set of experiments that SWEA$\oplus$OS is a highly effective model editing method. In addition, we show the results of SWEA$\oplus$OS and baselines performing 1, 10, 100, 1000, and 10,000 edits respectively on the \textsc{CounterFact} dataset in Appendix E.2. Additionally, in Appendix E.3, we present the qualitative results of the model generating facts after being edited on the \textsc{CounterFact} dataset.
\subsubsection{Sequential Batch Editing Results}
We use SWEA$\oplus$OS, PMET, MEMIT, and ROME to perform sequential batch editing on the GPT-J model in the \textsc{CounterFact} dataset. The number of sequences are 100, 20, 5, 2 with corresponding batch sizes of 1, 10, 100, 1000, respectively. The results are shown in Figure \ref{fig:SWEAOS:seq_edit}, indicating that SWEA$\oplus$OS performs the most stable performance in sequential batch editing. The performance of SWEA$\oplus$OS in sequential batch editing only show a slight decline as the batch size increased. When the editing batch is 1 and 1000, the scores of SWEA are 93.22 and 93.01, respectively. In contrast, the performances of PMET, MEMIT, and ROME are very unstable. The score of PMET and ROME decreased by 43.08\% and 42.74\% from an editing batch 1 to an editing batch 1000, respectively.
\subsection{Experiments on \textsc{RippleEdits}}
\begin{table*}[ht]
\centering
\begin{tabular}{crrrrrrr|rrrrrrrr}
\toprule
 \textbf{Dataset}& \textbf{Editor} & \textbf{LG} & \textbf{CI} & \textbf{CII} & \textbf{SA}  & \textbf{RS} & \textbf{Avg.} &\textbf{LG} & \textbf{CI} & \textbf{CII} & \textbf{SA}  & \textbf{RS} & \textbf{Avg.}\\  \hline   \multicolumn{8}{c|}{GPT-J} &\multicolumn{6}{c}{Llama-2}  \\  \hline
\multirow{3}{*}{RANDOM} &ROME & 0.58 & 0.52 & 0.24 & \textbf{1.0}  & \underline{0.44} &\underline{0.56} & 0.57 & 0.41 & 0.29 & \textbf{1.0}   & 0.52 & 0.56 \\
  &MEMIT & 0.60 & \underline{0.47} & 0.25 & 0.84  & \textbf{0.48} &0.53 &\textbf{0.67}  & 0.37 & \underline{0.33} &  \underline{0.89}  & \textbf{0.67} & \underline{0.59} \\
   &PMET & \textbf{0.70} & 0.46 & \underline{0.26} & \underline{0.88}  & 0.34 &0.53&\underline{0.62}  & \underline{0.47} & 0.18 &  \textbf{1.0}  & {0.49} & 0.55  \\
   & SWEA$\oplus$OS & \underline{0.62} & \textbf{0.54} & \textbf{0.63} & \textbf{1.0}   & 0.41 & \textbf{0.64} & 0.60 &\textbf{ 0.49 }& \textbf{0.37 }&  \textbf{1.0}  & \underline{0.55} & \textbf{0.60}\\ \hline
\multirow{3}{*}{POPULAR} & ROME & 0.30 & \underline{0.53} & \underline{0.28} & 0.86   & \underline{0.30 }&0.45&  \underline{0.28}&  0.39&  \underline{0.15}&    0.71&  0.32& 0.37  \\
   & MEMIT & 0.30 & 0.44 &0.19  & \textbf{1.0} &    \textbf{0.33}&0.45& \underline{0.28} &0.45  & 0.09 &  \textbf{0.96}  & \textbf{0.56} & \textbf{0.47} \\
      & PMET & \textbf{0.37} & 0.51 &0.17  & \underline{0.94} &    {0.29}&\underline{0.46} & \textbf{0.30} &\underline{0.47}  & 0.13 &  \underline{0.83}  & {0.31} & {0.41}\\
  & SWEA$\oplus$OS & \underline{0.32} & \textbf{0.56} & \textbf{0.53} & \textbf{1.0}   & 0.29 &\textbf{0.54}& \textbf{0.30} &  \textbf{0.49}& \textbf{0.16} &  0.81 &  \underline{0.37}&\underline{0.43}\\
  \bottomrule
\end{tabular}
\caption{Accuracy of \textsc{RippleEdits} on GPT-J and Llama-2.}\label{tab:SWEAOS:ripple}
\end{table*}
\begin{table*}[t]
\centering
\begin{tabular}{cccllllllll}
  \toprule
 \textbf{LLMs}&\textbf{Dataset}& \textbf{Editor} & \textbf{LG} & \textbf{CI} & \textbf{CII} & \textbf{SA}  & \textbf{RS} & \textbf{Avg.} \\  \hline
   \multirow{4}{*}{GPT-J} & \multirow{2}{*}{RANDOM}
   & SWEA$\oplus$OS & {0.62} & {0.54} & {0.63} & 1.0   & 0.41 & {0.64} \\
 &   &w/o suppressing & 0.60$_{\downarrow 0.02}$& 0.47$_{\downarrow 0.07}$ & 0.25$_{\downarrow 0.38}$ & 0.84$_{\downarrow 0.16}$ & {0.48}$_{\uparrow 0.07}$ &0.53$_{\downarrow 0.11}$ \\\cline{2-9}
 &  \multirow{2}{*}{POPULAR}
  & SWEA$\oplus$OS & {0.60} & {0.53} & {0.23} & {1.0}   & 0.38 &{0.54} \\
 &   &w/o suppressing  & 0.32$_{\downarrow 0.28}$ & 0.53 &0.19$_{\downarrow 0.04}$  & {0.86}$_{\downarrow 0.14}$ &    {0.28}$_{\downarrow 0.10}$&0.44 $_{\downarrow 0.1}$\\
 \hline
\multirow{4}{*}{Llama-2} &  \multirow{2}{*}{RANDOM}
   &SWEA$\oplus$OS  & 0.60 &{ 0.49 }& {0.37 }&  {1.0}  & 0.55 & {0.60} \\
  & &w/o suppressing  & 0.59 $_{\downarrow 0.01}$ & 0.49&0.30$_{\downarrow 0.07}$  & {1.0} &    {0.54}$_{\downarrow 0.01}$&0.58 $_{\downarrow 0.02}$\\\cline{2-9}
 & \multirow{2}{*}{POPULAR}
   &SWEA$\oplus$OS  & {0.30} &  {0.49}& {0.16} &  0.81  &  0.37&0.43 \\&&w/o suppressing  & 0.29 $_{\downarrow 0.01}$ & 0.49&0.16 & {0.81}&    {0.37}&0.42 $_{\downarrow 0.01}$ \\
  \bottomrule
\end{tabular}
\caption{Accuracy of \textsc{RippleEdits} on GPT-J and Llama-2 in ablation study.}\label{tab:SWEAOS:ablation}
\end{table*}
Since \textsc{RippleEdits} tests the model's ability to reason using edited knowledge, we first need to ensure that the model itself has the corresponding reasoning ability. Essentially, we need to edit the facts known to the model. To ensure this, each LLMs dataset needs to be filtered before \textsc{RippleEdits} testing. We followed the filtering steps of \textsc{RippleEdits}, finally generating 2188 and 2186 editing instances for GPT-J and Llama-2, respectively. Since none of these editing instances contain data for testing Preservation, our results do not include the Preservation metric.
\subsubsection{Results}
The accuracy of editing GPT-J and Llama-2 on the RANDOM and POPULAR subdatasets of \textsc{RippleEdits} is shown in Table \ref{tab:SWEAOS:ripple}. The results of GPT-J indicate that SWEA$\oplus$OS performs better than the baselines on CI, CII, and SA, suggesting that SWEA$\oplus$OS's ability to reason about edited knowledge surpasses existing baselines. For Llama-2, except for the LG and RS on the RANDOM dataset and the SA and RS on the POPULAR dataset where SWEA$\oplus$OS lags behind existing baselines, the results on other metrics are better than that of current baselines. From Table \ref{tab:SWEAOS:ripple}, it can be easily seen that SWEA$\oplus$OS performs poorly on the RS, which tests the ability of the editing method to retain non-edited knowledge about the subject. A possible reason for this situation is that SWEA$\oplus$OS introduced unintended knowledge during the optimization of the editing objectives. The metrics LG, CI and CII test the ability of the model editing method in 2-hop reasoning, and experimental results indicate that SWEAOS exhibits the best reasoning capability.
\subsection{Ablation Study}
To verify that the suppressing step in the OS fusion method is effective to the expression of new knowledge, we remove the suppressing step and test the results on the \textsc{RippleEdits} benchmark. As shown in Table \ref{tab:SWEAOS:ablation}, after removing the suppression step (i.e., w/o suppressing), the overwhelming majority of performance of SWEA$\oplus$OS in editing GPT-J and Llama-2 has declined, which indicates that our suppression step effectively alleviated the effect brought by KEDs of subject word embeddings. When editing GPT-J on the RANDOM and POPULAR datasets, the absence of suppression steps led to an average reduction of 0.14 in all metrics. Moreover, in the ablation experiment, the performance drop of GPT-J is more significant than that of Llama-2, which may be due to the stronger robustness brought by more parameters of Llama-2. Overall, the introduction of suppression steps in SWEA$\oplus$OS effectively facilitated the expression of new knowledge in LLMs.
\section{Conclusion}
We propose SWEA$\oplus$OS method for more effective and efficient knowledge editing. SWEA$\oplus$OS consists of Subject Word Embedding Fusion (SWEA) framework and the optimizing then suppressing (OS) fusion method. The SWEA framework uses token-level matching to identify the edited subject and adds the editing embeddings obtained from the OS fusion method to the subject embedding, ultimately altering the specific attributes of the subject to achieve knowledge editing. The OS fusion method employs an optimizing then suppressing strategy to effectively express new knowledge in editing embeddings.
SWEA$\oplus$OS achieve overall state-of-the-art (SOTA) results on the \textsc{CounterFact} and zsRE datasets, and it also shows SOTA performance in terms of reasoning ability on a more complex model editing benchmark \textsc{RippleEdits}.
Moreover, SWEA$\oplus$OS also provide a new insight to efficiently update knowledge of LLMs. Our code is available at \url{https://github.com/xpq-tech/SWEA.git}.

\section*{Ethical Statement}
The purpose of this work is to provide a more efficient and effective approach for knowledge editing. While SWEA can correct incorrect or outdated knowledge in LLMs cooperating with different fusion methods, it is important to recognize that SWEA is also susceptible to misuse, leading to the corruption of correct and already aligned knowledge in LLMs. Given that LLMs can inherently produce hallucinations, we would remind readers not to overly trust LLMs.
\section*{Acknowledgments}
This work was partly supported by the Hunan Provincial Natural Science Foundation Projects (No. 2022JJ30668 and No. 2022JJ30046), and also partly supported by the National Key R\&D Program of China (No. 2024YFB4506200).

\bibliography{refs}
\appendix
\section{Token-level Matching and Embedding Altering Algorithm}\label{alg:SWEAOS:key_match}
The token-level matching and embedding altering algorithm is showed in Algorithm. \ref{alg:SWEAOS:key_match}.
\begin{algorithm}[t]
\caption{Token-level Matching and Embedding Altering Algorithm}
\begin{algorithmic}
\STATE \textbf{Input:} Input batch token ids $b\_tids$, input representations $r$, and editing embeddings $\mathcal{E}$
\IF{$\mathcal{E}$ is empty}
    \RETURN
\ENDIF
\FOR{$i \gets 0$ \TO $b\_tids.\text{shape}[0]$}
    \IF{$b\_tids[i] > 1$}
        \STATE $tids \gets b\_tids[i]$, $max\_len\_key \gets \{\text{'key': None}\}$
        \FOR{$k \gets 0$ \TO $tids.size(0)$}
            \STATE $cur\_key \gets$ `'
            \FOR{$j \gets k$ \TO $tids.size(0)$}
                \STATE $cur\_key \gets cur\_key + \text{str}(tids[j].\text{item()}) + \_$
                \IF{$cur\_key[-1]$ exists in $\mathcal{E}$}
                    \IF{$\text{len}(cur\_key[-1]) > \text{len}(max\_len\_key[\text{'key'}])$}
                        \STATE $max\_len\_key[\text{'key'}] \gets cur\_key[:-1]$
                        \STATE $max\_len\_key[\text{'start'}] \gets k$
                        \STATE $max\_len\_key[\text{'end'}] \gets j+1$
                    \ENDIF
                \ENDIF
            \ENDFOR
        \ENDFOR
        \IF{$max\_len\_key[\text{'key'}] \neq \text{None}$}
            \STATE $r[i, max\_len\_key[\text{'start'}]: max\_len\_key[\text{'end'}], :] += \mathcal{E}[max\_len\_key[\text{'key'}]]$
        \ENDIF
    \ENDIF
\ENDFOR
\end{algorithmic}
\end{algorithm}
\section{Metircs}\label{appd:metric}
We use six metrics from previous work \cite{li2023pmet,meng2022massediting} on \textsc{CounterFact} and four metrics on zsRE to compare SWEA$\oplus$OS with the baselines. All metrics are designed such that a higher score is better.
\begin{itemize}
  \item {Efficacy}: This measures the rate at which the editing method successfully changes the knowledge of LLMs from $(\mathcal{T},\mathcal{Y})$ to $(\mathcal{T},\mathcal{Y}')$.
  \item {Generalization}: This evaluates the success rate of the post-edit model on $(\mathcal{T}^*,\mathcal{Y}')$, where $\mathcal{T}^*$ is the paraphrase of $\mathcal{T}$.
  \item {Specificity}: This assesses the success rate of the post-edit model on knowledge that hasn't been edited.
  \item {Score}: The harmonic mean of {Efficacy}, {Generalization}, and {Specificity}. It provides a comprehensive evaluation of the editing method.
  \item {Fluency}: This measures the generating power of the post-edit model on prompts containing the edited subject. This metric is only included in the \textsc{CounterFact} dataset.
  \item {Consistency}: This evaluates the degree of semantic consistency between the generated content of the post-edit model and $(\mathcal{T},\mathcal{Y}')$. This metric is only included in the \textsc{CounterFact} dataset.
\end{itemize}

\textsc{RippleEdits} mainly evaluates the ability of the editing method to capture the ripple effects of the edited fact, primarily using prompts within a 2-hop distance for evaluation. The evaluation metrics include:
\begin{itemize}
  \item {Logical Generalization (LG)}: This measures the success rate of the post-edit model on knowledge of facts that are logically symmetrical or transitive to the edited fact.
  \item {Compositionality \RNum{1} (C\RNum{1})}: This evaluates the ability of the editing method to combine 2-hop relationships starting with the subject.
  \item {Compositionality \RNum{2} (C\RNum{2})}: This evaluates the ability of the editing method to combine 2-hop relationships with the subject in the middle.
  \item {Subject Aliasing (SA)}: This assesses the generalization capability of the editing method on subject aliases.
  \item {Relation Specificity (RS)}: This measures the protection capability of the editing method on unedited facts on the subject.
\end{itemize}
\section{Implementation Details}\label{appendix:SWEAOS_impl_details}
We conducted all our experiments on the A800 GPU.
\subsection{Baselines}
Our baseline experiments on GPT-J(6B) follow the same setup as in prior work \cite{meng2022massediting,li2023pmet}. FT-W optimizes on layer 21 of Llama-2 with a learning rate of $5\times10^{-4}$ for a maximum of 25 steps. For ROME, MEMIT, and PMET, we use the EasyEdit configuration\footnote{https://github.com/zjunlp/EasyEdit} to edit the Llama-2, with ROME editing the 5th layer and MEMIT and PMET editing the \{4, 5, 6, 7, 8\} layers. The optimization proceeds for 20 steps with weight decay of 0.5, KL factor of 0.0625, and learning rate of $5\times10^{-1}$. The calculation of Covariance statistics is consistent with the setup of GPT-J(6B).

\subsection{SWEA$\oplus$OS}
The hyperparameters that need to be set for SWEA$\oplus$OS include the number of optimization steps, learning rate, weight decay, $\alpha$, $\beta$, and the clamping factor of $e$ during the optimization process. For the \textsc{CounterFact} and zsRE datasets, we set the optimization steps to 25, the learning rate to $2\times10^{-2}$, the weight decay to 0.3, $\alpha=0.2$, $\beta=1$. For the \textsc{RippleEdits} benchmark, we set the optimization steps for Llama2(7B) to 40, and $\alpha=0.3$ to ensure that the SWEA$\oplus$OS can optimize to meet the editing objectives, with other parameters consistent with the settings on the \textsc{CounterFact} and zsRE datasets. In all our experiments, we set the clamp factor to 1 and the Riemann integration steps to $n=20$ , and we retain dimensions with attribution scores more than $t=0.35$ times the maximum as KEDs. We set $\gamma=0.5$ to achieve the desired suppression effect while avoiding excessive suppression.
\begin{figure*}[t]
    \centering
\includegraphics[width=0.8\linewidth]{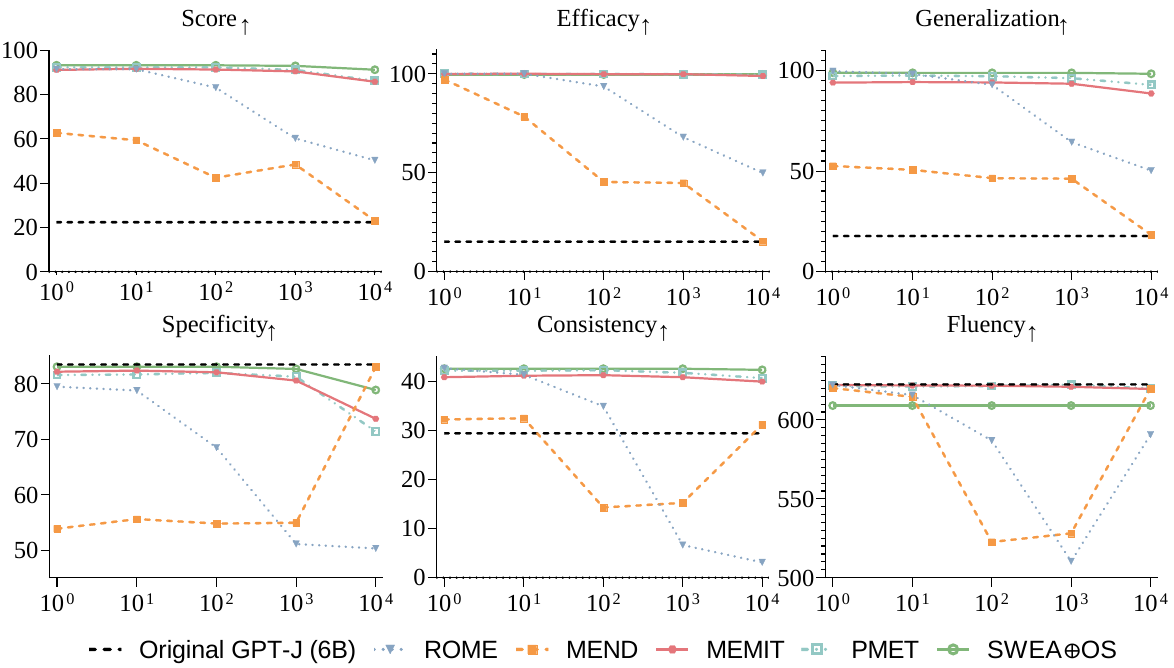}
    \caption{Editing results of SWEA$\oplus$OS and baselines with the number of edits (X-axis) on \textsc{COUNTERFACT} dataset.}\label{fig:SWEAOS:scale_up}
\end{figure*}
\section{Efficiency Analysis}\label{appendix:efficiency}
\begin{table}[t]
\centering
\begin{tabular}{cccc}
  \toprule
 \textbf{LLMs}&\textbf{SWEA$\oplus$OS}& \textbf{MEMIT} & \textbf{PMET} \\  \hline
  GPT-J & \textbf{14.86} &28.46 &40.34 \\  \hline
  Llama-2 &\textbf{27.46} &33.31 &48.84 \\
  \bottomrule
\end{tabular}
\caption{Execution time for 10,000 edits from scratch by SWEA$\oplus$OS, MEMIT, and PMET.}\label{tab:SWEAOS:exe_time}
\end{table}
\begin{figure}[t]
    \centering
\includegraphics[width=0.88\linewidth]{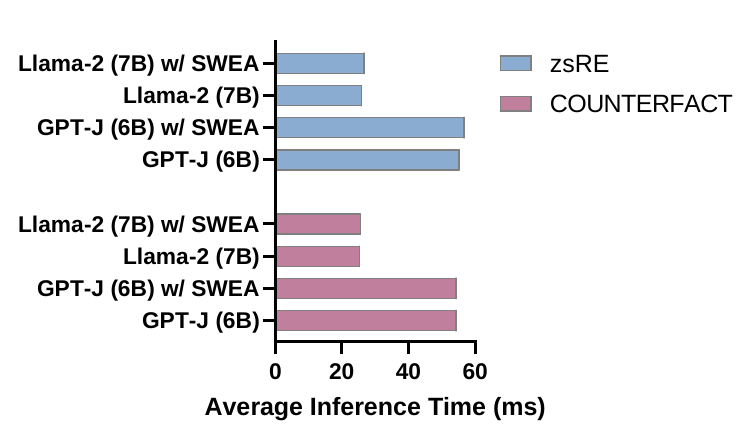}
    \caption{The average inference time of the original model and the model under the SWEA framework on the \textsc{CounterFact} and zsRE editing instances.}
    \label{fig:inferencetime}
\end{figure}
\subsection{Statistics of Execution Time}
In Table \ref{tab:SWEAOS:exe_time}, we present the time taken by SWEA, MEMIT, and PMET to perform 10,000 batch edits from scratch. The results in the table indicate that SWEA$\oplus$OS consumed the least amount of time, reducing the duration by 47.8\% and 17.6\% on GPT-J and Llama-2 respectively compared to MEMIT, and by 63.8\% and 43.8\% on GPT-J and Llama-2 respectively compared to PMET. This demonstrates the higher efficiency of our method. Note that local editing methods require prior causal tracing \cite{Meng2022Locating,meng2022massediting} to identify the critical layers used for editing, which also takes some time. We do not include this time in the execution time, which means that local editing methods actually take more time in practice.
\subsection{Inference Time Analysis}
The token-level matching of SWEA occurs on the CPU. Although it does not increase the inference load, it inevitably prolongs the inference time. To investigate this, we test the inference time of the original model and the model under the SWEA framework separately using editing instances from the \textsc{CounterFact} and zsRE datasets as inputs. The results, as shown in Figure \ref{fig:inferencetime}, indicate a slight increase in inference time for the model under the SWEA framework. Such millisecond-level delays are generally acceptable.
\section{Additional Results}
\subsection{Hyperparameter Analysis}
In this section, we analyze the two hyperparameters, $\gamma$ and $t$. The hyperparameter $\gamma$ controls the strength of suppression during the editing process, with larger values leading to stronger suppression. The hyperparameter $t$ is used to control the number of KEDs identified in the embedding. 

We first set different values of $\gamma$ to edit GPT-J on the RippleEdit dataset and explore the impact of $\gamma$ on editing performance. The results are shown in Table \ref{tab:diff_gamma}, which indicate that setting $\gamma$ either too high or too low adversely affects the editing performance, leading to a reduction in overall effectiveness. Next, we set different values of $t$ to edit GPT-J on the CounterFact dataset. The results, presented in Table \ref{tab:diff_t}, show that $t$ has a slight impact on editing performance, with the best performance observed at $t = 0.15$. Although our chosen value of $t = 0.35$ is not the optimal one in this case, it still achieves state-of-the-art performance. However, the performance is the worst when $t = 0$, suggesting that $t$ also needs to be set to an appropriate value.
\begin{table}[t]
\centering
\begin{tabular}{ccccccc}
\toprule
$\gamma$ & LG   & CI   & CII  & SA   & RS   & Avg.  \\ \hline
0        & 0.60 & 0.47 & 0.25 & 0.84 & 0.48 & 0.53  \\ \hline
0.25     & 0.60 & 0.52 & 0.19 & 1.0  & 0.35 & 0.53 \\ \hline
0.5      & 0.62 & 0.54 & 0.63 & 1.0  & 0.41 & 0.64  \\ \hline
0.75     & 0.61 & 0.52 & 0.15 & 1.0  & 0.78 & 0.61 \\ \hline
1.0      & 0.62 & 0.54 & 0.19 & 0.94 & 0.36 & 0.53  \\ \bottomrule
\end{tabular}
\caption{The impact of different $\gamma$ settings on SWEA$\oplus$OS when editing GPT-J in RippleEdits.}
\label{tab:diff_gamma}
\end{table}

\begin{table*}[ht]
\centering
\begin{tabular}{ccccccc}
\toprule
$t$    & Score & Efficacy & Generalization & Specificity & Fluency & Consistency \\ \hline
0.0    & 93.0 & 99.8     & 98.3           & 82.8       & 606.0   & 42.1       \\ \hline
0.15   & 93.1 & 99.8     & 98.7           & 82.8        & 610.6  & 42.6       \\ \hline
0.25   & 92.9 & 99.5     & 98.2           & 82.83       & 610.9  & 42.4       \\ \hline
0.35   & 93.0 & 99.8     & 98.3          & 82.8       & 609.3  & 42.2       \\ \hline
0.45   & 93.0 & 99.7     & 98.7          & 82.8       & 611.1  & 42.5       \\ \hline
0.55   & 93.1 & 99.6     & 98.7           & 82.9       & 611.5  & 42.7       \\ \bottomrule
\end{tabular}
\caption{The impact of different $t$ settings on SWEA$\oplus$OS when editing GPT-J in CounterFact.}
\label{tab:diff_t}
\end{table*}

\subsection{Results of Batch Scaling up}\label{appendix:scale_up}
We show in Figure \ref{fig:SWEAOS:scale_up} the results of SWEA$\oplus$OS and baselines performing 1, 10, 100, 1000, and 10,000 edits respectively on the \textsc{CounterFact} dataset. It can be seen that all editing methods have some decline in performance when expanding batch editing. SWEA$\oplus$OS performs more stably than the baselines, which highlights the stronger batch editing ability of SWEA$\oplus$OS.
\subsection{Qualitative Results}\label{appendix:QualitativeRes}
\begin{figure*}[t]
    \centering
    \begin{subfigure}[b]{0.9\textwidth}
        \centering
        \includegraphics[width=\textwidth]{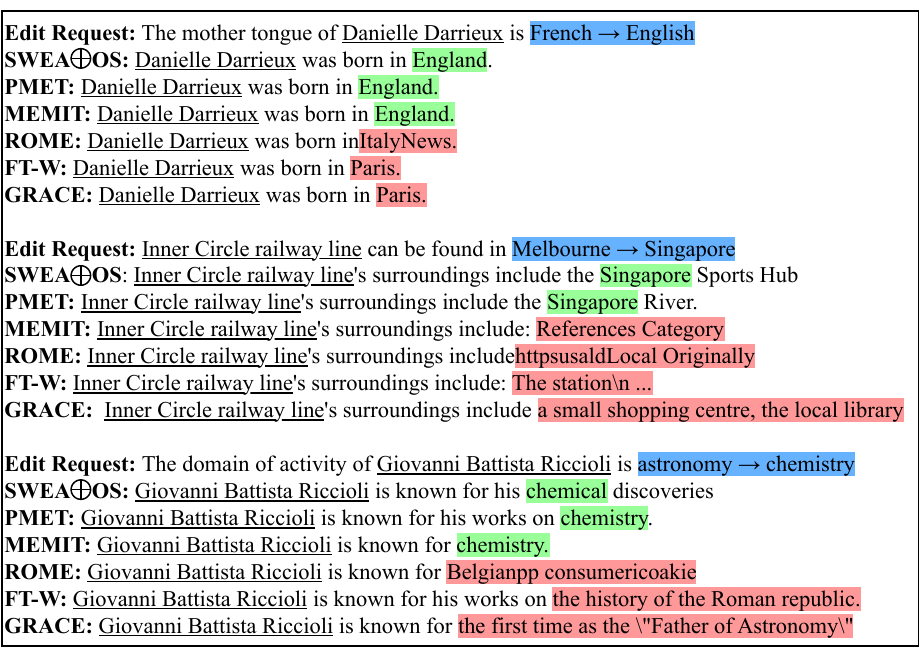} 
        \caption{GPT-J}
        \label{fig:qr-gptj}
    \end{subfigure}
    \hfill
    \begin{subfigure}[b]{0.9\textwidth}
        \centering
        \includegraphics[width=\textwidth]{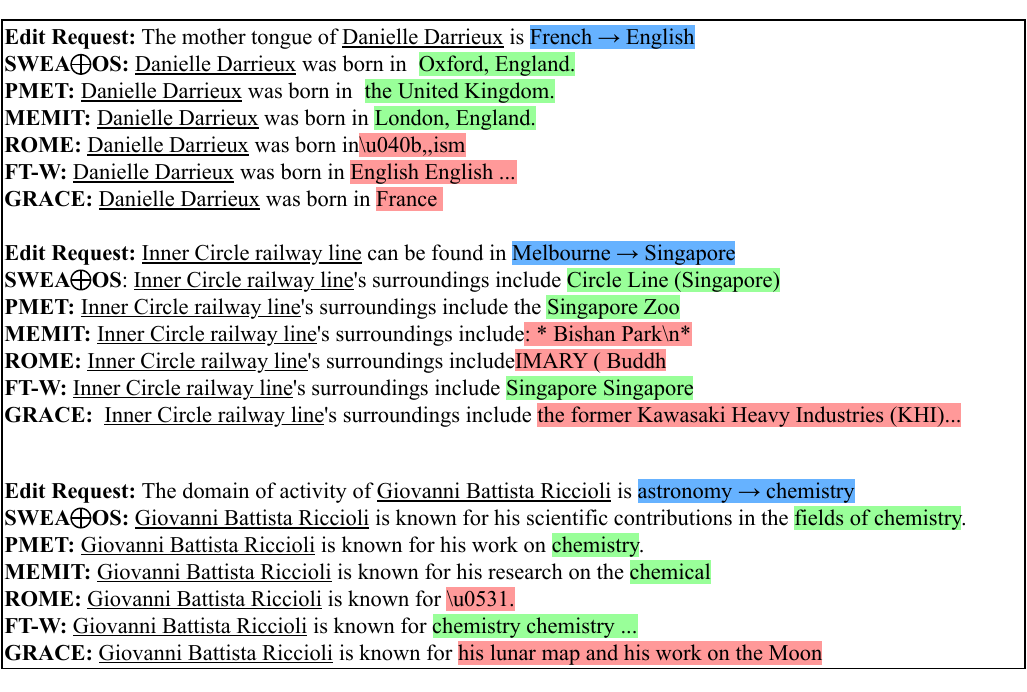}  
        \caption{Llama2}
        \label{fig:qr-llama2}
    \end{subfigure}
    \caption{Qualitative results in \textsc{CounterFact} dataset. The underlined text represents the subject. The blue background indicates the editing requirement; the green background signifies that the generated content is logically consistent with the required facts, while the red background indicates inconsistency or unrelated.}
    \label{fig:qr}
\end{figure*}
In Figures \ref{fig:qr-gptj} and \ref{fig:qr-llama2}, we present some qualitative results of editing GPT-J and LLaMA2 on the \textsc{CounterFact} dataset under the setting of an editing batch of 10,000 for each editing method. These results show that models edited by SWEAOS, PMET, and MEMIT can generate content consistent with the required facts, indicating that these editing methods may have allowed the models to internalize the knowledge to some extent. In contrast, GPT-J edited by FT-W fails to generate content consistent with the required facts, while Llama-2 edited by FT can do so but tends to produce repetitive content. ROME and GRACE perform the worst, with the ROME-edited model tending to generate disordered content, and the GRACE-edited model generating content unrelated to the required facts.
\end{document}